\documentclass{article} % For LaTeX2e
\usepackage{iclr2027_conference,times}

\usepackage{amsmath,amsfonts,bm}

\def\eqref#1{equation~\ref{#1}}
\def\1{\bm{1}}

\DeclareMathAlphabet{\mathsfit}{\encodingdefault}{\sfdefault}{m}{sl}
\SetMathAlphabet{\mathsfit}{bold}{\encodingdefault}{\sfdefault}{bx}{n}

\usepackage{algorithm}
\usepackage{algpseudocode}
\usepackage{graphicx} 
\usepackage{hyperref}
\usepackage{url}
\usepackage{subcaption}
\usepackage{booktabs}
\usepackage{makecell}
\usepackage{multirow}
\title{Evolving Neural Solvers for Multi-objective Combinatorial Optimization}

\author{
Ya-Hui Jia$^{1}$,  Le-Xin~Wen$^{1}$,
Yang Wang$^{1}$\thanks{Corresponding author.},
Wei-Neng Chen$^{2}$,
Yi Mei$^{3}$\\
$^{1}$School of Future Technology, South China University of Technology, China\\
$^{2}$School of Computer Science and Engineering, South China University of Technology, China\\
$^{3}$School of Engineering and Computer Science, Victoria University of Wellington, New Zealand\\
\texttt{jia.yahui@foxmail.com,saulwen369@gmail.com,}\\
\texttt{ftwangyang@mail.scut.edu.cn, cwnraul634@aliyun.com, yi.mei@vuw.ac.nz}
}

\begin{document}

\maketitle
\begin{abstract}
Multi-objective combinatorial optimization (MOCO) seeks high-quality solutions that represent different trade-offs among conflicting objectives. Existing neural methods typically train a dedicated model for each preference or condition a single model on different preferences. Training dedicated models becomes more costly as the number of preferences increases, while a shared model can produce duplicate solutions for different preferences, limiting the benefit of denser preference queries. We propose the Evolving Neural Solver (ENS), which trains only a small set of preference-specific models and takes advantage of evolutionary computation to evolve them into a diverse collection of solvers for better solution quality and Pareto front coverage. ENS first trains a model at an estimated knee-point preference to target a balanced compromise, then obtains the remaining initial models through parameter transfer toward the extremes, so as to preliminarily approximate the Pareto front. It then evolves these models to find more candidates that can generate non-dominated solutions through evolution strategies. Experimental results on classic MOCO problems show that ENS outperforms state-of-the-art neural MOCO methods in terms of hypervolume in most evaluated settings.
\end{abstract}

\section{Introduction}

Many combinatorial optimization problems involve conflicting objectives. In vehicle routing, for example, minimizing total travel distance may conflict with minimizing the longest vehicle route. Similar trade-offs arise in scheduling and knapsack problems \citep{zajac2021objectives,turkyilmaz2020research,ishibuchi2014behavior}. Multi-objective combinatorial optimization (MOCO) therefore seeks Pareto-optimal solutions, each representing a trade-off that cannot be improved in one objective without worsening another. Since computing the complete Pareto set is often intractable, the practical goal is to obtain a high-quality approximation that represents a broad range of trade-offs.

Neural approaches to MOCO follow two main directions. Single-model methods condition one shared network on the preference vector, so that arbitrarily many preferences can be queried after a single training run \citep{lin2022pareto,wang2023multiobjective,fan2024conditional,chen2025rethinking}. A single policy, however, has to cover the entire preference simplex, and neighboring preferences frequently decode to identical solutions \citep{chen2023neural}, so querying more preferences does not necessarily enrich the resulting front. Multi-model methods avoid this ambiguity by assigning a dedicated policy to each preference \citep{li2020deep,zhang2022meta,chen2023efficient}, but every such policy has to be obtained through reinforcement learning, which converges slowly and makes the acquisition of each additional solver inefficient.

Considering the strength of population-based search in evolutionary computation, we propose the Evolving Neural Solver (ENS), a two-stage framework that combines reinforcement learning with evolutionary computation. ENS trains only a small set of preference-specific models, termed \emph{anchors}, and then expands them into a diverse collection of solvers. In the first stage, since the objectives of a MOCO problem differ in optimization difficulty and in magnitude, ENS starts from the estimated knee-point preference and trains the first anchor there. The remaining anchors are then obtained by successive transfer toward the extremes. This small set of anchors already provides a coarse approximation of the Pareto front. In the second stage, the anchors serve as the starting points of a population-based search, in which new solvers are repeatedly generated around the existing ones and retained only when they contribute a solution that is non-dominated among all candidates, thereby enriching the Pareto front.

The contributions of this paper are summarized as follows:

\begin{itemize}
\item We propose ENS, which expands a small set of preference-specific neural models through evolutionary search. Joint non-dominated selection across populations retains solvers that contribute different trade-offs, extending the solution set without separate training for every preference.

\item We introduce an anchor placement and training strategy that starts from an estimated knee-point preference and transfers successively toward the extremes. Anchoring the limited training budget at the most critical region of the preference space makes the fullest use of each trained model and provides stronger starting points for the subsequent evolutionary search.
\end{itemize}

Experiments on classic MOCO problems show that the proposed framework outperforms state-of-the-art neural MOCO methods in terms of hypervolume in most evaluated settings.

\section{Related Work}

\paragraph{Conventional MOCO methods.}
Conventional MOCO methods include exact algorithms and heuristics. Exact algorithms provide optimality guarantees, but computing the complete Pareto set can become prohibitively expensive as problem size grows \citep{ehrgott2005multicriteria}. Evolutionary frameworks approximate the Pareto front through population search. NSGA-II uses non-dominated sorting and crowding distance \citep{deb2002fast}, whereas MOEA/D coordinates scalarized subproblems \citep{zhang2007moea}. SPEA2 and PAES use archive-based mechanisms \citep{zitzler2001spea2,knowles2000approximating}, while NSGA-III and RVEA maintain diversity using reference points or vectors \citep{fan2014evolutionary,cheng2016reference}. MOEA/D-DE and GDE3 incorporate differential evolution \citep{li2008multiobjective,kukkonen2005gde3}, and MOPSO employs particle-swarm search \citep{coello2004handling}. These general-purpose frameworks can be adapted to discrete problems through suitable representations and operators. Genetic and Pareto local search further exploit problem structure \citep{jaszkiewicz2002genetic,shi2022improving}. Although effective, these methods generally require substantial search for each new instance, motivating learned policies that reuse experience across instances.

\paragraph{Neural MOCO methods.}
Neural combinatorial optimization was initially developed around single-objective problems. Pointer Networks learn sequential solution construction \citep{vinyals2015pointer}, while the attention model of \citet{kool2018attention} uses Transformer-based representations \citep{vaswani2017attention} and reinforcement learning. POMO improves training through multiple starting points and a shared reward baseline \citep{kwon2020pomo}. Subsequent work improves representations and generalization \citep{wang2025distance,gao2023towards,zhou2026instance,meng2025eformer,meng2026uniteformer}, with further advances in routing policies and multi-task learning \citep{li2023learning,zhou2024mvmoe,liao2025bopo}.

For MOCO, single-model methods share one network across preferences. \citet{lin2022pareto} uses a preference-conditioned hypernetwork to generate decoder parameters, whereas \citet{wang2023multiobjective} directly conditions the policy on objective weights. \citet{fan2024conditional} extends conditioning to support generalization across preferences and problem sizes. More recent methods improve preference representations through weight embeddings \citep{chen2025rethinking} and enhance instance representations through graph-image multimodal fusion \citep{chen2025neural}. POCCO introduces conditional computation and pairwise preference optimization \citep{fan2026preference}, while WeCon strengthens weight conditioning in both the encoder and decoder \citep{wu2026wecon}. Multi-model methods instead obtain specialized policies for different preferences. \citet{li2020deep} combines scalarization with parameter transfer between neighboring preferences, while meta-learning methods learn initializations that support efficient adaptation \citep{zhang2022meta,chen2023efficient}. Neural evolutionary methods also optimize learned policies: \citet{zhang2021modrl} evolves model parameters and retains a fixed-size population using non-dominated sorting and crowding distance, while \citet{shao2021multi} combines neural learning with evolutionary optimization for better performance.

\section{Preliminaries}

\subsection{Multi-objective Combinatorial Optimization}

\paragraph{Problem Definition.}
A MOCO problem is defined by a discrete feasible set $\mathcal{X}$, whose elements are combinatorial structures such as permutations or subsets induced by the problem constraints, together with $m\geq 2$ objective functions $f_i:\mathcal{X}\to\mathbb{R}$. It minimizes the objective vector:
\begin{equation}
\min_{x\in\mathcal{X}}\mathbf{f}(x)
=
\bigl(f_1(x),\ldots,f_m(x)\bigr).
\end{equation}

\paragraph{Pareto Dominance and Optimality.}
A solution $x_1$ dominates $x_2$, denoted by $x_1\prec x_2$, if $f_i(x_1)\leq f_i(x_2)$ for all $i$ and the inequality is strict for at least one objective. A feasible solution is Pareto-optimal if it is not dominated by any feasible solution. The set of Pareto-optimal solutions is the Pareto set $\mathcal{X}^*$, and its image $\{\mathbf{f}(x):x\in\mathcal{X}^*\}$ is the Pareto front.

\paragraph{Extreme Points and Knee Point.}
For each objective $i$, let $x^{(i)}$ be a Pareto-optimal solution among the minimizers of $f_i$ alone,
$x^{(i)} \in \operatorname*{arg\,min}_{x\in\mathcal{X}} f_i(x), x^{(i)}\in\mathcal{X}^*$, and call $\mathbf{f}(x^{(i)})$ the $i$-th \emph{extreme point} of the Pareto front, namely the objective vector attained when objective $i$ alone is optimized. Let $H$ be the hyperplane through the $m$ extreme points and $\hat{\mathbf{n}}$ its unit normal pointing in the direction of decreasing objectives. Following \citet{das1996normal,das1999characterizing}, the signed distance from $H$ and the knee solution $x^\dagger$ are defined as:
\begin{equation}
b(x)=\hat{\mathbf{n}}^{\top}\bigl(\mathbf{f}(x)-\mathbf{p}\bigr),
\qquad
x^\dagger\in\operatorname*{arg\,max}_{x\in\mathcal{X}^*} b(x),
\end{equation}
where $\mathbf{p}\in H$ is arbitrary and $b(x)$ is independent of its choice. The objective vector $\mathbf{f}(x^\dagger)$ is the \emph{knee point}, the Pareto-optimal trade-off that improves most upon the linear interpolation of the $m$ extreme points.

\subsection{Preference Scalarization and Neural Construction}

For a preference vector $\boldsymbol{\lambda}\in\Delta^{m-1}$, where $\Delta^{m-1}=\{\boldsymbol{\lambda}\in\mathbb{R}_{\geq0}^{m}:\sum_{i=1}^{m}\lambda_i=1\}$, weighted-sum scalarization minimizes $\sum_{i=1}^{m}\lambda_i f_i(x)$. A common alternative is Tchebycheff scalarization, which minimizes $\max_i\lambda_i|f_i(x)-z_i^*|$. For strictly positive weights, any global optimum of the weighted-sum problem is Pareto-optimal, although weighted sums do not necessarily recover the entire Pareto front.

For a fixed preference, a neural constructive solver builds a solution $x=(a_1,\ldots,a_T)$ sequentially for an instance $s$:
\begin{equation}
\pi_\theta(x\mid s)
=
\prod_{t=1}^{T}
\pi_\theta(a_t\mid a_{1:t-1},s),
\end{equation}
where $a_t$ is the action at step $t$ and $T$ is the number of construction steps. Under weighted-sum scalarization, $\pi_\theta$ is trained to maximize $\mathbb{E}_{s\sim\mathcal{D},\,x\sim\pi_\theta(\cdot\mid s)}[-\sum_{i=1}^{m}\lambda_i f_i(x)]$. Thus, serving $J$ target preferences with dedicated policies requires training or fine-tuning $J$ models.

\section{Methodology}

\begin{figure}[t]
    \centering
    \includegraphics[width=\textwidth]{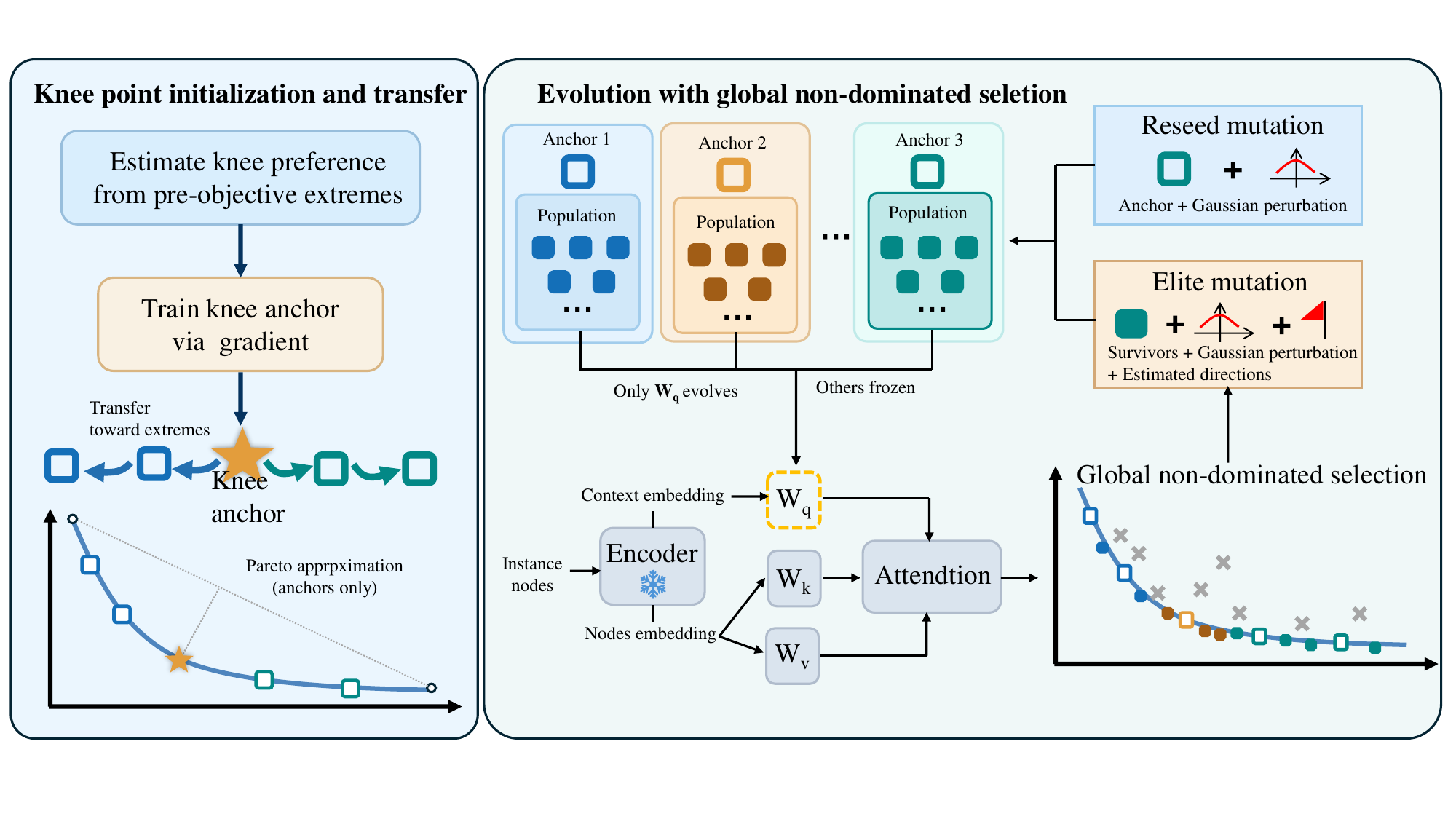}
    \caption{Overview of ENS}
    \label{figure1}
\end{figure}

\subsection{Overview and Solver Backbone}

Figure~\ref{figure1} summarizes the ENS pipeline. ENS proceeds in two stages. In the first stage, ENS estimates the knee-point preference of the problem, trains one anchor there from scratch, and obtains the remaining $M-1$ anchors by successive transfer toward the extreme preferences, so that a small number of trained models already cover the preference space coarsely. In the second stage, each anchor becomes the starting point of a population that evolves in a low-dimensional subspace of its decoder. Candidates from all anchors are evaluated jointly, and only those contributing a non-dominated solution are retained, which extends the initial $M$ anchors into a much larger collection of solvers. At inference, every retained solver is rolled out on the test instance, and the non-dominated solutions among all produced trajectories form the final Pareto front approximation, without requiring any preference as input.

Every solver in ENS shares the same encoder-decoder backbone adopted from POMO \citep{kwon2020pomo}, and they differ only in their parameters. The encoder maps an instance $s$ to node embeddings $h_1,\dots,h_n\in\mathbb{R}^d$, which depend only on the instance and are computed once per rollout. The decoder holds three linear projections $W_q,W_k,W_v\in\mathbb{R}^{d\times d}$. $W_k$ and $W_v$ are applied once to all node embeddings and cached, while $W_q$ is applied at every decoding step $t$ to the context embedding $h_c$ formed from the current decoding state:
\begin{equation}
q_t = W_q\,h_c, \qquad k_i = W_k\,h_i,\qquad v_i=W_v\,h_i .
\end{equation}
Multi-head attention over $(q_t,k,v)$, followed by a learned output projection, is then dotted with the node embeddings to give the next-node probabilities:
\begin{equation}
u_{t,i} = C\cdot\tanh\!\Big(\frac{\mathrm{MHA}(q_t,k,v)^\top h_i}{\sqrt{d}}\Big),\qquad \pi_\theta(a_t{=}i\mid a_{1:t-1},s)=\frac{\exp(u_{t,i})}{\sum_{j}\exp(u_{t,j})} ,
\end{equation}
where $C$ is the tanh clipping constant.
\subsection{Stage One: Knee-Point Initialization and Transfer}\label{sec:stage1}

\paragraph{Locating the knee point.}
Estimating the knee-point preference requires the per-objective extreme points $\mathbf{f}(x^{(i)})$. At an extreme preference the MOCO problem reduces to a single-objective problem, which we solve with a strong heuristic or exact method for the problem at hand on a small number of sampled instances, averaging the resulting objective vectors into one representative extreme point per objective (per-problem details are given in Appendix~\ref{app:knee_point}). Let $\mathbf{z}^*$ denote the ideal point, $z_i^*=f_i(x^{(i)})$, and let $\mathbf{e}=(1,\ldots,1)^\top$. The extreme points and the ideal point form the payoff matrix:
\begin{equation}
\Phi =
\begin{pmatrix}
f_1(x^{(1)})-z_1^* & f_1(x^{(2)})-z_1^* & \cdots & f_1(x^{(m)})-z_1^* \\
f_2(x^{(1)})-z_2^* & f_2(x^{(2)})-z_2^* & \cdots & f_2(x^{(m)})-z_2^* \\
\vdots & \vdots & \ddots & \vdots \\
f_m(x^{(1)})-z_m^* & f_m(x^{(2)})-z_m^* & \cdots & f_m(x^{(m)})-z_m^*
\end{pmatrix},
\end{equation}
whose $i$-th column $\mathbf{f}(x^{(i)})-\mathbf{z}^*$ records how far the $i$-th extreme solution stands from the ideal point on every objective. \citet{das1999characterizing} shows that the preference vector $\boldsymbol\lambda^\dagger$ associated with the knee point is the normalized solution of the $m\times m$ linear system:
\begin{equation}
\Phi^\top \boldsymbol\omega = \mathbf{e},
\qquad
\boldsymbol\lambda^\dagger = \boldsymbol\omega \big/ \textstyle\sum_{i=1}^{m}\omega_i .
\end{equation}

\paragraph{Anchor placement.}
All anchors are positioned relative to $\boldsymbol\lambda^\dagger$, so that none of them sits at a fixed geometric location. For a bi-objective problem, the segment between $\boldsymbol\lambda^\dagger$ and each extreme preference is divided into $(M+1)/2$ equal sub-intervals, where $M$ is the number of anchors. The sub-interval adjacent to $\boldsymbol\lambda^\dagger$ is absorbed into the region that the knee anchor itself already represents, and each of the remaining sub-intervals receives one anchor at its midpoint. Implementation details are described in Appendix~\ref{app:knee_point}.

\paragraph{Transfer training.}
Under a fixed preference $\boldsymbol\lambda$, the weighted-sum scalarized cost of a solution is $g(x\mid\boldsymbol\lambda)=\boldsymbol\lambda^\top\mathbf{f}(x)$, and an anchor with parameters $\theta$ is trained to minimize its expectation,
\begin{equation}
\mathcal{L}(\theta\mid\boldsymbol\lambda)=\mathbb{E}_{s\sim\mathcal{D}}\,\mathbb{E}_{x\sim\pi_\theta(\cdot\mid s)}\bigl[g(x\mid\boldsymbol\lambda)\bigr],
\end{equation}
with REINFORCE following the POMO training paradigm \citep{kwon2020pomo}, in which a rollout starts from $N$ different first nodes and the resulting $N$ trajectories serve as baselines for one another. Only the knee anchor is trained from a random initialization for $E_0$ epochs,
\begin{equation}
\theta^\dagger = \operatorname*{arg\,min}_{\theta}\ \mathcal{L}(\theta\mid\boldsymbol\lambda^\dagger).
\end{equation}
The remaining anchors are then trained one at a time, outwards from the knee point. Each anchor $c$ is warm-started from the anchor nearest to it among those already trained, $\theta^{(c)}\leftarrow\theta^{(c')}$, and adapted to its own preference $\boldsymbol\lambda^{(c)}$ for $E_1$ epochs, so that in the bi-objective case the anchors propagate along a chain on either side of the knee anchor. Such a warm start converges in far fewer epochs than training every anchor from scratch. 

\subsection{Stage Two: Evolutionary Extension}

The number of parameters in a full solver is far beyond an evolution algorithm can search efficiently, since its sample cost grows rapidly with the dimension of the search space. The subspace that is evolved should therefore be as small as possible while still being the part that determines a solver's trade-off. In a complete solver, the encoder extracts the geometric and demand features of an instance, a representation that is only weakly related to the objective weighting, so the encoder is kept frozen throughout this stage. Within the decoder, a change of preference amounts to a change in what kind of node the current state should attend to, which in the attention mechanism corresponds directly to the semantic space of the query, $W_q$. The other two projections play a different role: $W_k$ maps the node embeddings into the key space and mainly performs a fixed feature alignment, while $W_v$ only aggregates information once the attention weights have been determined. We therefore take $W_q$ as the individual that undergoes evolution, and keep all other weights frozen. The evolutionary extension repeats three steps over $G$ generations: solution generation, solution evaluation, and searching direction update. Algorithm~\ref{alg:ens} in Appendix~\ref{app:algorithm} gives the complete procedure.

\paragraph{Solution generation.}
We flatten $W_q\in\mathbb{R}^{d\times d}$ into a vector $\mathbf{w}\in\mathbb{R}^{D}$ with $D=d^2$, which serves as an individual of the evolutionary population, while the encoder and all other decoder weights are inherited unchanged from the corresponding anchor. Starting from the $M$ anchor vectors $\{\mathbf{w}_c\}_{c=1}^{M}$, each anchor maintains its own population. Offspring are generated in one of two ways, which play different roles. Reseed offspring mutate the anchor vector itself and keep exploring broadly around the anchor:
\begin{equation}
\mathbf{w}'_{\text{reseed}} = \mathbf{w}_c + \boldsymbol\epsilon, \qquad \boldsymbol\epsilon \sim \mathcal{N}(\mathbf{0}, \sigma^2 I),
\end{equation}
whereas elite offspring mutate a survivor of the previous generation, drawn cyclically from $S_c$, so as to further refine solvers that have already proven competitive:
\begin{equation}
\mathbf{w}'_{\text{elite}} = \mathbf{w}_{\text{parent}} + \boldsymbol\epsilon + \alpha\,\boldsymbol\delta_i^{c}, \qquad \mathbf{w}_{\text{parent}} \in S_c,\ \boldsymbol\epsilon \sim \mathcal{N}(\mathbf{0}, \sigma^2 I),\ i \in \{1,\dots,m\},
\end{equation}
where elite offspring are split evenly among the $m$ search directions $\boldsymbol\delta_1^{c},\dots,\boldsymbol\delta_m^{c}$ maintained by anchor $c$. In the first generation there are no survivors yet, so all offspring are generated from the anchor; from the second generation onward, offspring are allocated between the two ways in the ratio $r:(1-r)$. Together with the survivors, they form the candidate pool $P_c$ of the current generation.

\paragraph{Solution evaluation.}
A candidate is not scored by a single fitness value, but by a small set of representative points. A discrete set of target preferences is fixed in advance, and each of them is assigned to its nearest anchor, so that every anchor $c$ owns a set $\Lambda_c$ of preferences for which its candidates are held responsible. Each candidate $\mathbf{w}'\in P_c$ is reshaped back into $W_q$, substituted into the otherwise frozen backbone, and rolled out on the same batch of $U$ instances, which returns $N$ trajectories per instance.

One representative point is then computed for each preference $\boldsymbol\lambda\in\Lambda_c$ in two steps. On every instance, the $N$ trajectories may have mutually non-dominated objective vectors, so one of them is selected as the trajectory with the lowest scalarized cost $g(x\mid\boldsymbol\lambda)$, a step we call the \emph{internal selection}. The objective vectors of the $U$ selected trajectories, one per instance, are then averaged, which gives the representative point $\bar{\mathbf{f}}(\mathbf{w}'\mid\boldsymbol\lambda)\in\mathbb{R}^{m}$ and reduces the evaluation noise of a single instance. Repeating this over $\Lambda_c$ describes the candidate by $|\Lambda_c|$ representative points, which summarize how it performs across the region of the preference space its anchor is responsible for.

Selection is then made over all candidates at once. A candidate may possess from 1 to $|\Lambda_c|$ distinct representative points. The representative points of every candidate are pooled into a single set. A candidate survives into the next generation if it has at least one globally non-dominated representative point. Finally, each survivor of anchor $c$ is assigned to the group $\mathcal{B}_i^{c}$ of whichever objective $i$ its non-dominated points improve upon the anchor on, and a survivor may belong to several groups.

\paragraph{Searching direction update.}
The groups $\mathcal{B}_i^{c}$ provide the signal for updating the search directions, which follow the search-gradient estimator of natural evolution strategies \citep{wierstra2014natural,schaul2011high}. For a group indicator $\mathbb{1}(\boldsymbol\epsilon)\in\{0,1\}$, marking whether an offspring perturbed by $\boldsymbol\epsilon$ falls into the group, the gradient of the expected group membership with respect to the parent vector is:
\begin{equation}
\nabla_{\mathbf{w}}\,\mathbb{E}_{\boldsymbol\epsilon\sim\mathcal{N}(\mathbf{0},\sigma^2 I)}\bigl[\mathbb{1}(\boldsymbol\epsilon)\bigr] = \frac{1}{\sigma^2}\,\mathbb{E}\bigl[\mathbb{1}(\boldsymbol\epsilon)\,\boldsymbol\epsilon\bigr],
\end{equation}
which is estimated from the current generation by the mean perturbation of the offspring in that group. Each direction is then refreshed by an exponential moving average:
\begin{equation}
\boldsymbol\delta_i^{c} \leftarrow \beta\,\boldsymbol\delta_i^{c} + (1-\beta)\,\frac{\bar{\boldsymbol\epsilon}_i^{c}}{\lVert\bar{\boldsymbol\epsilon}_i^{c}\rVert},
\qquad
\bar{\boldsymbol\epsilon}_i^{c} = \frac{1}{|\mathcal{B}_i^{c}|}\sum_{k \in \mathcal{B}_i^{c}} \boldsymbol\epsilon_k ,
\end{equation}
so that the elite offspring of the next generation are pushed towards the region where survivors have recently been found on that side of the anchor. After $G$ generations, each anchor's surviving population is merged with its own vector, giving the final set of solvers.

\section{Experiment}

\subsection{Experiment Settings}\label{sec:5.1}

\paragraph{Problems.} We evaluate the multi-objective traveling salesman problem (MOTSP) \citep{lust2010multiobjective} in its bi-objective (Bi-TSP) and tri-objective (Tri-TSP) forms, the multi-objective capacitated vehicle routing problem (MOCVRP) \citep{zajac2021objectives} in its bi-objective form (Bi-CVRP), and the multi-objective knapsack problem (MOKP) \citep{ishibuchi2014behavior} in its bi-objective form (Bi-KP); full problem definitions appear in Appendix~\ref{app:problem_def}. Bi-TSP, Bi-CVRP, and Tri-TSP are each tested at sizes $n=20/50/100$, and Bi-KP at $n=50/100/200$.

\paragraph{Hyperparameters.}
Most of our hyperparameters follow prior work \citep{kwon2020pomo,lin2022pareto}. For the bi-objective problems, we use $M=5$ anchors, except for Bi-TSP100, Bi-CVRP100 and Bi-KP200, where $M=7$ anchors are used; for Tri-TSP we use $M=16$ anchors, except for Tri-TSP100, where $M=28$. The knee anchor is trained from scratch for $E_0=200$ epochs, each epoch drawing 100,000 randomly sampled instances, while every other anchor is warm-started from its nearest trained neighbor and adapted for $E_1=100$ epochs when $M=5$, or $E_1=50$ epochs otherwise. All anchors are optimized with Adam \citep{kinga2015method} using a learning rate of $10^{-4}$ and a weight decay of $10^{-6}$, and a rollout starts from $N=n$ different first nodes following POMO. In the evolutionary stage, each anchor generates $K=1{,}000$ offspring per generation, with a reseed ratio of $r=0.5$, an EMA rate of $\beta=0.9$, and evaluates them on a batch of $U=100$ sampled instances, for a total of $G=100$ generations.

\paragraph{Baselines.} We compare ENS against three groups of representative methods. The single-model group consists of \textbf{PMOCO} \citep{lin2022pareto} and the weight-embedding method \textbf{WE} \citep{chen2025rethinking}, for which we report both of its variants, \textbf{WE-Add} and \textbf{WE-CA}. The multi-model group consists of \textbf{DRL-MOA} \citep{li2020deep}, \textbf{MDRL} \citep{zhang2022meta} and \textbf{EMNH} \citep{chen2023efficient}. The non-learnable group covers general-purpose evolutionary algorithms, \textbf{MOEA/D} \citep{zhang2007moea} and \textbf{NSGA-II} \citep{deb2002fast}, both run for 4,000 iterations; evolutionary algorithms tailored to MOCO, \textbf{MOGLS} \citep{jaszkiewicz2002genetic}, run for 4,000 iterations with 100 local search steps per iteration, and \textbf{PPLS/D-C} \citep{shi2022improving}, run for 200 generations; \textbf{WS-LKH} and \textbf{WS-DP}, which combine weighted-sum scalarization with the strong LKH heuristic \citep{helsgaun2000effective,tinos2018efficient} for MOTSP and with a dynamic programming solver for MOKP. Two further single-model methods, \textbf{MORAM} \citep{wang2023multiobjective} and the size-generalizing \textbf{CNH} \citep{fan2024conditional}, are included only in the generalization experiments of Appendix~\ref{app:Generalization experiments}. Every neural baseline is trained under the same weighted-sum scalarization as ENS to keep the comparison fair. All experiments are conducted on a server equipped with two NVIDIA RTX 4090 GPUs.

 \paragraph{Metrics.} We evaluate solution quality with the normalized hypervolume (HV) \citep{while2006faster} indicator, where a larger value indicates a Pareto front that is both closer to the true front and more diverse; its definition, together with the reference and normalization point used for each problem and size, is given in Appendix~\ref{app:hv}. For every method we report the average HV, its gap relative to the ENS variant on that problem, and the total time needed to solve 200 test instances. Results obtained with instance augmentation (Appendix~\ref{app:aug}), which we apply to every problem except MOKP, are marked with the suffix ``-Aug''. In all tables, the best and the second-best results in each column are highlighted in \textbf{bold} and \underline{underlined}, respectively, and tied results are marked in the same way.
 
\begin{table*}[p]
\centering
\caption{Results on Bi-TSP, Bi-CVRP, Bi-KP and Tri-TSP.}
\label{tab:main_results}
\scriptsize
\setlength{\tabcolsep}{2.2pt}
\renewcommand{\arraystretch}{0.85}
\begin{tabular*}{\textwidth}{@{\extracolsep{\fill}}c|lccccccccc@{}}
\toprule
& & \multicolumn{3}{c}{Bi-TSP20} & \multicolumn{3}{c}{Bi-TSP50} & \multicolumn{3}{c}{Bi-TSP100} \\
\cmidrule(lr){3-5} \cmidrule(lr){6-8} \cmidrule(lr){9-11}
& Method & HV$\uparrow$ & Gap$\downarrow$ & Time$\downarrow$ & HV$\uparrow$ & Gap$\downarrow$ & Time$\downarrow$ & HV$\uparrow$ & Gap$\downarrow$ & Time$\downarrow$ \\
\midrule
\multirow{5}{2.1cm}{\centering\tiny Non-learnable}
& WS-LKH   & 0.6270 & 0.44\% & 10m & \underline{0.6415} & \underline{0.31\%} & 1.8h & \textbf{0.7090} & \textbf{-0.24\%} & 6.0h \\
& MOEA/D   & 0.6241 & 0.91\% & 1.7h & 0.6316 & 1.85\% & 1.8h & 0.6899 & 2.46\% & 2.2h \\
& NSGA-II  & 0.6258 & 0.64\% & 6.0h & 0.6120 & 4.90\% & 6.1h & 0.6692 & 5.39\% & 6.9h \\
& MOGLS    & 0.6279 & 0.30\% & 1.6h & 0.6330 & 1.63\% & 3.7h & 0.6854 & 3.10\% & 11h  \\
& PPLS/D-C & 0.6256 & 0.67\% & 26m  & 0.6282 & 2.38\% & 2.8h & 0.6844 & 3.24\% & 11h  \\
\midrule
\multirow{3}{2.1cm}{\centering\tiny Single-model}
& PMOCO  & 0.6259 & 0.62\% & 6s & 0.6351 & 1.31\% & 10s & 0.6957 & 1.64\% & 19s \\
& WE-Add & 0.6270 & 0.44\% & 6s & 0.6386 & 0.76\% & 10s & 0.7024 & 0.69\% & 19s \\
& WE-CA  & 0.6270 & 0.44\% & 6s & 0.6391 & 0.68\% & 10s & 0.7039 & 0.48\% & 20s \\
\midrule
\multirow{4}{2.1cm}{\centering\tiny Multi-model}
& DRL-MOA & 0.6257 & 0.65\% & 6s & 0.6360 & 1.17\% & 9s & 0.6970 & 1.46\% & 21s \\
& MDRL    & 0.6271 & 0.43\% & 5s & 0.6364 & 1.10\% & 9s & 0.6969 & 1.47\% & 17s \\
& EMNH    & 0.6271 & 0.43\% & 5s & 0.6364 & 1.10\% & 9s & 0.6969 & 1.47\% & 16s \\
& ENS     & \underline{0.6292} & \underline{0.10\%} & 2s & 0.6392 & 0.67\% & 8s & 0.7029 & 0.62\% & 45s \\
\midrule
\multirow{3}{2.1cm}{\centering\tiny Single-model-Aug}
& PMOCO-Aug  & 0.6270 & 0.44\% & 1.1m & 0.6395 & 0.62\% & 3.2m & 0.7016 & 0.81\% & 15m \\
& WE-Add-Aug & 0.6272 & 0.41\% & 1.1m & 0.6411 & 0.37\% & 3.2m & 0.7058 & 0.21\% & 15m \\
& WE-CA-Aug  & 0.6271 & 0.43\% & 1.1m & 0.6412 & 0.36\% & 3.3m & 0.7067 & 0.08\% & 15m \\
\midrule
\multirow{3}{2.1cm}{\centering\tiny Multi-model-Aug}
& MDRL-Aug & 0.6271 & 0.43\% & 33s & 0.6408 & 0.42\% & 1.7m & 0.7022 & 0.72\% & 14m \\
& EMNH-Aug & 0.6271 & 0.43\% & 33s & 0.6408 & 0.42\% & 1.7m & 0.7023 & 0.71\% & 14m \\
& ENS-Aug  & \textbf{0.6298} & \textbf{0.00\%} & 1.8m & \textbf{0.6435} & \textbf{0.00\%} & 7.5m & \underline{0.7073} & \underline{0.00\%} & 36m \\
\midrule
& & \multicolumn{3}{c}{Bi-CVRP20} & \multicolumn{3}{c}{Bi-CVRP50} & \multicolumn{3}{c}{Bi-CVRP100} \\
\cmidrule(lr){3-5} \cmidrule(lr){6-8} \cmidrule(lr){9-11}
& Method & HV$\uparrow$ & Gap$\downarrow$ & Time$\downarrow$ & HV$\uparrow$ & Gap$\downarrow$ & Time$\downarrow$ & HV$\uparrow$ & Gap$\downarrow$ & Time$\downarrow$ \\
\midrule
\multirow{4}{2.1cm}{\centering\tiny Non-learnable}
& MOEA/D   & 0.4255 & 1.21\% & 2.3h & 0.4000 & 2.65\% & 2.9h & 0.3953 & 3.28\% & 5.0h \\
& NSGA-II  & 0.4275 & 0.74\% & 6.4h & 0.3896 & 5.18\% & 8.8h & 0.3620 & 11.43\% & 9.4h \\
& MOGLS    & 0.4278 & 0.67\% & 9.0h & 0.3984 & 3.04\% & 20h  & 0.3875 & 5.19\% & 72h  \\
& PPLS/D-C & 0.4287 & 0.46\% & 1.6h & 0.4007 & 2.48\% & 9.7h & 0.3946 & 3.45\% & 38h  \\
\midrule
\multirow{3}{2.1cm}{\centering\tiny Single-model}
& PMOCO  & 0.4267 & 0.93\% & 7s & 0.4036 & 1.78\% & 12s & 0.3913 & 4.26\% & 27s \\
& WE-Add & 0.4292 & 0.35\% & 6s & 0.4089 & 0.49\% & 12s & 0.4061 & 0.64\% & 26s \\
& WE-CA  & 0.4293 & 0.33\% & 6s & 0.4090 & 0.46\% & 12s & 0.4072 & 0.37\% & 25s \\
\midrule
\multirow{4}{2.1cm}{\centering\tiny Multi-model}
& DRL-MOA & 0.4287 & 0.46\% & 10s & 0.4076 & 0.80\% & 12s & 0.4055 & 0.78\% & 33s \\
& MDRL    & 0.4291 & 0.37\% & 8s  & 0.4082 & 0.66\% & 13s & 0.4056 & 0.76\% & 32s \\
& EMNH    & 0.4299 & 0.19\% & 7s  & 0.4098 & 0.27\% & 13s & 0.4072 & 0.37\% & 31s \\
& ENS     & 0.4297 & 0.23\% & 3s & 0.4097 & 0.29\% & 6s & 0.4078 & 0.22\% & 14s \\
\midrule
\multirow{3}{2.1cm}{\centering\tiny Single-model-Aug}
& PMOCO-Aug  & 0.4294 & 0.30\% & 14s & 0.4080 & 0.71\% & 36s & 0.3969 & 2.89\% & 2.7m \\
& WE-Add-Aug & 0.4300 & 0.16\% & 14s & 0.4103 & 0.15\% & 38s & 0.4079 & 0.20\% & 2.5m \\
& WE-CA-Aug  & 0.4301 & 0.14\% & 14s & 0.4104 & 0.12\% & 37s & \underline{0.4082} & \underline{0.12\%} & 2.5m \\
\midrule
\multirow{3}{2.1cm}{\centering\tiny Multi-model-Aug}
& MDRL-Aug & 0.4294 & 0.30\% & 11s & 0.4092 & 0.41\% & 36s & 0.4072 & 0.37\% & 2.8m \\
& EMNH-Aug & \underline{0.4302} & \underline{0.12\%} & 11s & \underline{0.4106} & \underline{0.07\%} & 35s & 0.4079 & 0.20\% & 2.8m \\
& ENS-Aug  & \textbf{0.4307} & \textbf{0.00\%} & 8s & \textbf{0.4109} & \textbf{0.00\%} & 34s & \textbf{0.4087} & \textbf{0.00\%} & 1.8m \\
\midrule
& & \multicolumn{3}{c}{Bi-KP50} & \multicolumn{3}{c}{Bi-KP100} & \multicolumn{3}{c}{Bi-KP200} \\
\cmidrule(lr){3-5} \cmidrule(lr){6-8} \cmidrule(lr){9-11}
& Method & HV$\uparrow$ & Gap$\downarrow$ & Time$\downarrow$ & HV$\uparrow$ & Gap$\downarrow$ & Time$\downarrow$ & HV$\uparrow$ & Gap$\downarrow$ & Time$\downarrow$ \\
\midrule
\multirow{5}{2.1cm}{\centering\tiny Non-learnable}
& WS-DP    & \underline{0.3561} & \underline{0.14\%} & 22m & 0.4532 & 0.11\% & 2.0h & \underline{0.3601} & \underline{0.00\%} & 5.8h \\
& MOEA/D   & 0.3540 & 0.73\% & 1.6h & 0.4508 & 0.64\% & 1.7h & 0.3581 & 0.56\% & 1.8h \\
& NSGA-II  & 0.3547 & 0.53\% & 7.8h & 0.4520 & 0.37\% & 8.0h & 0.3590 & 0.31\% & 8.4h \\
& MOGLS    & 0.3540 & 0.73\% & 5.8h & 0.4510 & 0.60\% & 10h  & 0.3582 & 0.53\% & 18h  \\
& PPLS/D-C & 0.3528 & 1.07\% & 18m  & 0.4480 & 1.26\% & 47m  & 0.3541 & 1.67\% & 1.5h \\
\midrule
\multirow{3}{2.1cm}{\centering\tiny Single-model}
& PMOCO  & 0.3552 & 0.39\% & 9s & 0.4523 & 0.31\% & 19s & 0.3595 & 0.17\% & 1.3m \\
& WE-Add & 0.3558 & 0.22\% & 8s & 0.4530 & 0.15\% & 20s & 0.3600 & 0.03\% & 1.1m \\
& WE-CA  & \underline{0.3561} & \underline{0.14\%} & 9s & 0.4533 & 0.09\% & 19s & \underline{0.3601} & \underline{0.00\%} & 1.1m \\
\midrule
\multirow{4}{2.1cm}{\centering\tiny Multi-model}
& DRL-MOA & 0.3559 & 0.20\% & 9s & 0.4531 & 0.13\% & 18s & \underline{0.3601} & \underline{0.00\%} & 1.0m \\
& MDRL    & 0.3530 & 1.01\% & 6s & 0.4532 & 0.11\% & 21s & \underline{0.3601} & \underline{0.00\%} & 1.2m \\
& EMNH    & \underline{0.3561} & \underline{0.14\%} & 6s & \underline{0.4535} & \underline{0.04\%} & 21s & \textbf{0.3603} & \textbf{-0.06\%} & 1.2m \\
& ENS     & \textbf{0.3566} & \textbf{0.00\%} & 7s & \textbf{0.4537} & \textbf{0.00\%} & 44s & \underline{0.3601} & \underline{0.00\%} & 3m \\
\midrule
& & \multicolumn{3}{c}{Tri-TSP20} & \multicolumn{3}{c}{Tri-TSP50} & \multicolumn{3}{c}{Tri-TSP100} \\
\cmidrule(lr){3-5} \cmidrule(lr){6-8} \cmidrule(lr){9-11}
& Method & HV$\uparrow$ & Gap$\downarrow$ & Time$\downarrow$ & HV$\uparrow$ & Gap$\downarrow$ & Time$\downarrow$ & HV$\uparrow$ & Gap$\downarrow$ & Time$\downarrow$ \\
\midrule
\multirow{5}{2.1cm}{\centering\tiny Non-learnable}
& WS-LKH   & 0.4712 & 1.40\% & 12m & \underline{0.4440} & \underline{1.92\%} & 1.9h & \underline{0.5076} & \underline{0.90\%} & 6.6h \\
& MOEA/D   & 0.4702 & 1.61\% & 1.9h & 0.4314 & 4.71\% & 2.2h & 0.4511 & 11.93\% & 2.4h \\
& NSGA-II  & 0.4238 & 11.32\% & 7.1h & 0.2858 & 36.87\% & 7.5h & 0.2824 & 44.87\% & 9.0h \\
& MOGLS    & 0.4701 & 1.63\% & 1.5h & 0.4211 & 6.98\% & 4.1h & 0.4254 & 16.95\% & 13h  \\
& PPLS/D-C & 0.4698 & 1.69\% & 1.4h & 0.4174 & 7.80\% & 3.9h & 0.4376 & 14.56\% & 14h  \\
\midrule
\multirow{3}{2.1cm}{\centering\tiny Single-model}
& PMOCO  & 0.4693 & 1.80\% & 5s & 0.4315 & 4.68\% & 8s & 0.4858 & 5.15\% & 18s \\
& WE-Add & 0.4705 & 1.55\% & 6s & 0.4379 & 3.27\% & 10s & 0.4943 & 3.49\% & 20s \\
& WE-CA  & 0.4707 & 1.51\% & 5s & 0.4389 & 3.05\% & 9s & 0.4976 & 2.85\% & 19s \\
\midrule
\multirow{4}{2.1cm}{\centering\tiny Multi-model}
& DRL-MOA & 0.4699 & 1.67\% & 6s & 0.4303 & 4.95\% & 9s & 0.4806 & 6.17\% & 19s \\
& MDRL    & 0.4699 & 1.67\% & 5s & 0.4317 & 4.64\% & 9s & 0.4852 & 5.27\% & 16s \\
& EMNH    & 0.4699 & 1.67\% & 5s & 0.4324 & 4.48\% & 9s & 0.4866 & 5.00\% & 16s \\
& ENS     & \underline{0.4719} & \underline{1.26\%} & 5s & 0.4317 & 4.64\% & 8s & 0.4937 & 3.61\% & 27s \\
\midrule
\multirow{3}{2.1cm}{\centering\tiny Single-model-Aug}
& PMOCO-Aug  & 0.4712 & 1.40\% & 5.1m & 0.4409 & 2.61\% & 28m & 0.4956 & 3.24\% & 1.7h \\
& WE-Add-Aug & 0.4712 & 1.40\% & 5.1m & 0.4429 & 2.16\% & 29m & 0.5016 & 2.07\% & 1.8h \\
& WE-CA-Aug  & 0.4712 & 1.40\% & 5.2m & 0.4431 & 2.12\% & 29m & 0.5036 & 1.68\% & 1.7h \\
\midrule
\multirow{3}{2.1cm}{\centering\tiny Multi-model-Aug}
& MDRL-Aug & 0.4712 & 1.40\% & 2.6m & 0.4408 & 2.63\% & 25m & 0.4958 & 3.20\% & 1.7h \\
& EMNH-Aug & 0.4712 & 1.40\% & 2.6m & 0.4418 & 2.41\% & 25m & 0.4973 & 2.91\% & 1.7h \\
& ENS-Aug  & \textbf{0.4779} & \textbf{0.00\%} & 4.6m & \textbf{0.4527} & \textbf{0.00\%} & 25m & \textbf{0.5122} & \textbf{0.00\%} & 3.3h \\
\bottomrule
\end{tabular*}
\end{table*}

\subsection{Results and Analyses}\label{sec:5.2}

Table~\ref{tab:main_results} reports results on Bi-TSP, Bi-CVRP, Bi-KP and Tri-TSP. ENS attains the best HV on ten of the twelve problem-size combinations, the two exceptions being Bi-TSP100, where the near-exact WS-LKH remains $0.24\%$ ahead, and Bi-KP200, where EMNH leads by $0.0002$. On Bi-TSP100 ENS-Aug still surpasses every neural baseline, augmented or not, and on Bi-TSP20 ENS alone (0.6292) already exceeds every augmented baseline, including WE-Add-Aug (0.6272). On Bi-CVRP, ENS-Aug leads the strongest single-model baseline WE-CA-Aug at every size, and on Bi-KP, ENS alone surpasses every baseline at $n=50$ and $n=100$.

The margin of ENS widens with the number of objectives. On Tri-TSP, ENS-Aug attains the best hypervolume at every size, leading the strongest neural baseline by $1.40\%$, $2.12\%$ and $1.68\%$ at $n=20/50/100$, and it is the only learning-based method that surpasses WS-LKH on Tri-TSP100 ($0.5122$ against $0.5076$). With three objectives, the Pareto front is a surface rather than a curve, so a fixed set of preferences covers it far more thinly than in the bi-objective case; the evolutionary stage, which admits any candidate that contributes a non-dominated solution, is not tied to such a set and therefore benefits the most exactly where the front is hardest to cover.

Generalization to larger sizes and to real-world instances is reported in Appendix~\ref{app:Generalization experiments}.

\subsection{Ablation Study}\label{sec:5.3}

\paragraph{Effect of evolution and of the evolved weights.}
Table~\ref{tab:ablation_evolution_weights} reports the ablation results on Bi-TSP50. Without the evolutionary stage, the trained anchors alone lose $2.13\%$ of hypervolume, more than twice the largest gap among the choices of evolved weights. Evolving $W_q$ is the best choice, and evolving all three projections jointly is no better, although it triples the dimension of the search space: the query projection is the smallest subspace that carries the decision preference, so enlarging the search beyond it costs more in sample efficiency than the added expressiveness returns.

\begin{figure}[h]
    \centering
    \begin{subfigure}[b]{0.32\textwidth}
        \centering
        \includegraphics[width=\textwidth]{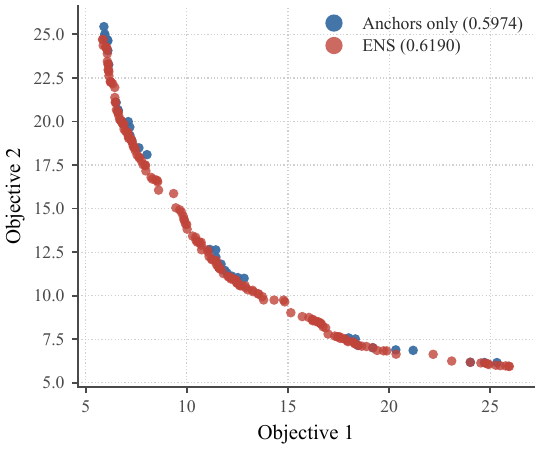}
        \caption{Effect of evolution}
        \label{fig:ablation_evolution}
    \end{subfigure}
    \hfill
    \begin{subfigure}[b]{0.32\textwidth}
        \centering
        \includegraphics[width=\textwidth]{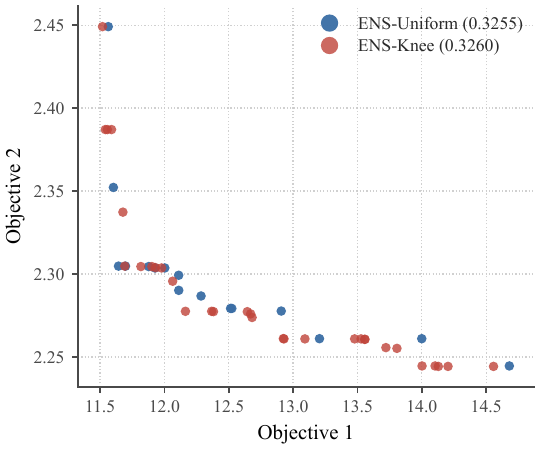}
        \caption{Effect of anchor placement}
        \label{fig:ablation_placement}
    \end{subfigure}
    \hfill
    \begin{subfigure}[b]{0.32\textwidth}
        \centering
        \includegraphics[width=\textwidth]{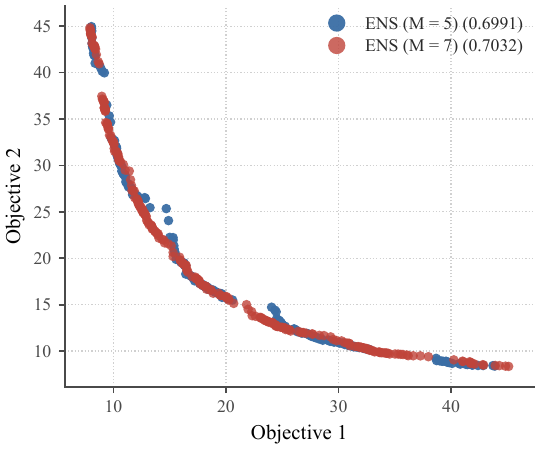}
        \caption{Effect of the number of anchors}
        \label{fig:ablation_anchors}
    \end{subfigure}
    \caption{Pareto fronts obtained on a single test instance. (a) Bi-TSP50, the anchors alone against the full ENS; (b) Bi-CVRP50, anchors placed uniformly against anchors placed relative to the estimated knee point; (c) Bi-TSP100, $M=5$ against $M=7$ anchors.}
    \label{fig:ablation_fronts}
\end{figure}

\paragraph{Qualitative comparison of the resulting fronts.}
Figure~\ref{fig:ablation_fronts} shows the fronts obtained on one test instance of each ablation. In Figure~\ref{fig:ablation_evolution}, the anchors alone produce a sparse and fragmented front, with solutions clustered around the few preferences they were trained for, whereas evolution turns it into a dense and continuous one, which is the per-instance picture behind the $2.13\%$ gap above.

Figure~\ref{fig:ablation_placement} compares the two placement rules on Bi-CVRP50. Uniform anchors spread their budget evenly and leave a sparser front, thin in particular around the knee, which is where the front bends and contributes the most hypervolume per solution; knee-based placement concentrates the solutions exactly there.

Figure~\ref{fig:ablation_anchors} varies the number of anchors on Bi-TSP100. Since a single query projection has a limited expressive range, the region an anchor can reach through evolution is bounded, and on larger instances that range no longer suffices: with $M=5$ the front dents at the boundaries between neighboring anchors, and each dent costs hypervolume, while $M=7$ fills them. Anchors therefore do more than supply good starting points, as they also compete with one another through the global non-dominated selection, so denser anchors improve the quality of the front.

\begin{table}[h]
\centering
\footnotesize
\caption{Effect of evolution and of the evolved weights on Bi-TSP50.}
\label{tab:ablation_evolution_weights}
\begin{tabular}{lccc}
\toprule
Method & HV$\uparrow$ & Gap$\downarrow$ & Time$\downarrow$ \\
\midrule
Anchors only            & 0.6256 & 2.13\% & 3s \\
Evolve $W_q$ (ours)     & \textbf{0.6392} & \textbf{0.00\%} & 8s \\
Evolve $W_k$            & 0.6379 & 0.20\% & 7s \\
Evolve $W_v$            & 0.6335 & 0.89\% & 8s \\
Evolve $W_q,W_k,W_v$    & 0.6378 & 0.22\% & 7s \\
\bottomrule
\end{tabular}
\end{table}

\section{Conclusion}

The goal of this paper was to effectively solve MOCO problems with high-quality solutions and broad coverage of the Pareto front. This goal has been successfully achieved by proposing the Evolving Neural Solver (ENS), which combines reinforcement learning with evolutionary computation. Specifically, ENS first trains a small set of anchors, starting from an estimated knee-point preference and transferring successively toward the extremes, so that the limited training budget is concentrated on the most critical region of the preference space. It then evolves the decoder query projections of these anchors and retains every candidate that contributes a non-dominated solution, expanding a few trained models into a diverse collection of solvers. Comparisons with strong neural and conventional methods show that ENS achieves the best hypervolume in most evaluated settings, and on the three-objective TSP it is the only learning-based method that surpasses WS-LKH. Further analyzes verify each component's effectiveness and the evolved solvers' ability to generalize to larger sizes and real-world instances.

Beyond the problems studied here, we believe that the paradigm embodied by ENS, in which learning provides strong initial solvers and evolution extends them into a diverse population, can transfer to a broader range of combinatorial optimization problems and to other domains like intelligent control and multi-agent systems.

\section*{AI use statement}

We used a large language model to help polish the writing of this paper, including improving clarity and wording in the text.

% \section*{Reproducibility statement}

% All hyperparameters used for training the anchors and running the evolutionary stage are reported in Section~\ref{sec:5.1}, and the complete evolutionary extension algorithm is given as pseudocode in Appendix~\ref{app:algorithm}. Problem formulations, the normalized hypervolume definition with its reference and ideal points, and the procedure for locating the knee point are detailed in Appendices~\ref{app:problem_def}, \ref{app:hv}, and \ref{app:knee_point} respectively. Upon acceptance, we will publicly release our code, trained model weights, and test set.

\bibliography{reference}
\bibliographystyle{iclr2027_conference}

\clearpage
\appendix

\section{Problem Definition}\label{app:problem_def}
\paragraph{MOTSP.}
We consider the MOTSP with $n$ nodes, where each node carries $m$ pairs of 2D coordinates, one per objective. Let $L_i(x)$ denote the total length of tour $x$ measured under the $i$-th coordinate system. Given a preference $\boldsymbol{\lambda} = (\lambda_1, \ldots, \lambda_m) \in \Delta^{m-1}$, MOTSP is scalarized as minimizing the preference-weighted sum of tour lengths across all coordinate systems:
\begin{equation}
\min_{x} \sum_{i=1}^{m} \lambda_i L_i(x).
\end{equation}
\paragraph{MOCVRP.}
We consider the Bi-CVRP with $n$ customer nodes and one depot, served by a fleet of vehicles, each with capacity $Q$. Let $\mathrm{TD}(x)$ denote the total routing distance of solution $x$ and $\mathrm{MS}(x)$ denote its makespan, i.e., the length of the single longest route. Given a preference $\boldsymbol{\lambda} = (\lambda_1, \lambda_2) \in \Delta^1$, MOCVRP is scalarized as:
\begin{equation}
\min_{x} \ \lambda_1 \mathrm{TD}(x) + \lambda_2 \mathrm{MS}(x).
\end{equation}

\paragraph{MOKP.}
We consider the Bi-KP with $n$ items and a shared capacity $B$, where a solution $x$ selects a feasible subset of items with total weight at most $B$. Let $V_i(x)$ denote the total value of $x$ under the $i$-th value assignment, $i = 1, 2$. Given a preference $\boldsymbol{\lambda} = (\lambda_1, \lambda_2) \in \Delta^1$, MOKP is scalarized as:
\begin{equation}
\max_{x} \ \lambda_1 V_1(x) + \lambda_2 V_2(x).
\end{equation}

\section{Hypervolume Indicator}\label{app:hv}

HV is a standard indicator for evaluating a set of Pareto-optimal solutions, since it jointly measures both convergence to the true Pareto front and diversity along it.
Since HV is computed from a finite approximation $\mathcal{F}$ of the Pareto front, we further need a reference point $\mathbf{r}\in\mathbb{R}^m$, dominated by every point in $\mathcal{F}$, and normalization point $\mathbf{z}_0\in\mathbb{R}^m$ satisfying $z_{0,i} < \min\{f_i(x) : \mathbf{f}(x)\in\mathcal{F}\}$ for every $i$.
The hypervolume of $\mathcal{F}$ with respect to $\mathbf{r}$ is:
\begin{equation}
\mathrm{HV}_{\mathbf{r}}(\mathcal{F}) = \mu\Big(\bigcup_{\mathbf{f}(x)\in\mathcal{F}} [\mathbf{f}(x), \mathbf{r}]\Big), \qquad [\mathbf{f}(x), \mathbf{r}] = \prod_{i=1}^{m} [f_i(x), r_i],
\end{equation}
where $\mu$ is the Lebesgue measure and $[\mathbf{f}(x), \mathbf{r}]$ is the axis-aligned box between $\mathbf{f}(x)$ and $\mathbf{r}$: a larger union of these boxes means the solutions are both closer to the ideal region and more spread out along the front.
We report the normalized hypervolume:
\begin{equation}
\mathrm{HV}'_{\mathbf{r}}(\mathcal{F}) = \mathrm{HV}_{\mathbf{r}}(\mathcal{F}) \Big/ \prod_{i=1}^{m} |r_i - z_{0,i}|,
\end{equation}
so that a larger value indicates a Pareto front that is both closer to optimal and more diverse. For the maximization problem MOKP, the inequalities are reversed accordingly
All methods share the same $\mathbf{r}$ and $\mathbf{z}_0$ for a given problem and size, listed in Table~\ref{tab:hv_ref_ideal}.

\begin{table}[H]
\centering
\scriptsize
\caption{Reference points $\mathbf{r}$ and normalization points $\mathbf{z}_0$ used for hypervolume.}
\label{tab:hv_ref_ideal}
\begin{tabular}{llcc}
\toprule
Problem & $n$ & $\mathbf{r}$ & $\mathbf{z}_0$ \\
\midrule
\multirow{3}{*}{Bi-TSP}  & 20  & (20,20)&(0,0) \\
                         & 50  & (35,35) & (0,0) \\
                         & 100 & (65,65) & (0,0) \\

\midrule
\multirow{3}{*}{Bi-CVRP}    & 20  & (30,4) & (0,0) \\
                         & 50  & (45,4) &  (0,0)\\
                         & 100 & (80,4) & (0,0) \\
\midrule
\multirow{3}{*}{Bi-KP}      & 50  & (5,5) & (30,30) \\
                         & 100 & (20,20) & (50,50) \\
                         & 200 & (30,30) & (75,75) \\
\midrule
\multirow{3}{*}{Tri-TSP} & 20  & (20,20,20) & (0,0,0) \\
                         & 50  & (35,35,35) & (0,0,0)\\
                         & 100 & (65,65,65)& (0,0,0)\\
\bottomrule
\end{tabular}
\end{table}

\section{Instance Augmentation}\label{app:aug}

At inference time, we apply instance augmentation to improve solution quality: an instance is transformed into a set of equivalent instances that share the same optimal solution, each is solved independently, and the best solution across all transformed instances is retained.
For MOCVRP, an instance admits 8 transformations of its 2-dimensional coordinates, applied identically to the depot and every customer node, $\{(x,y),\,(y,x),\,(x,1-y),\,(y,1-x),\,(1-x,y),\,(1-y,x),\,(1-x,1-y),\,(1-y,1-x)\}$.

For an $m$-objective TSP instance, each of the $m$ groups of 2-dimensional coordinates independently admits the same 8 transformations, giving $8^m$ transformed instances in total.
This augmentation relies on the 2-dimensional Euclidean coordinates of MOTSP and MOCVRP, and does not apply to MOKP, which has no spatial coordinates.

\section{Anchor placement}\label{app:knee_point}
\subsection{Implementation Details of Locating Knee Point}
For every problem, the extreme points required by the payoff matrix $\Phi$ (Section \ref{sec:stage1}) are estimated by reducing the problem to the corresponding single objective, solving it on $N_0=10$ sampled instances of the target size with an existing, non-neural method appropriate to that problem, and averaging the resulting objective vectors into one representative point per extreme.

For MOTSP (Bi-TSP and Tri-TSP), each node carries two or three sets of 2D coordinates, one per objective; at extreme $i$, the reduced single-objective TSP uses only that objective's coordinate subspace, and is solved with LKH-3.

For MOCVRP, the two extremes reduce to (i) minimizing total route distance with no per-route length limit, and (ii) minimizing total distance subject to a per-route length cap set at the theoretical lower bound on the makespan, $\mathrm{MS}_0=2\max_c\,\mathrm{dist}(\mathrm{depot},c)$, which turns the second extreme into a makespan-oriented solution. Both are solved with a hybrid genetic search CVRP solver under a size-dependent wall-clock budget per instance ($5\mathrm{s}/15\mathrm{s}/30\mathrm{s}$ at $n=20/50/100$).

For MOKP, the two extremes use only one of the two item values as the knapsack objective and are solved exactly with OR-Tools' knapsack solver.

Table~\ref{tab:knee_points} reports the knee-point preference $\boldsymbol\lambda^\dagger$ estimated by our method for each problem and problem size. For Bi-TSP, Bi-KP, and Tri-TSP, the objectives are symmetric by construction, so the estimated knee point stays close to an even split. Bi-CVRP behaves differently: its two objectives, the total travel distance and the single-vehicle makespan, differ substantially in raw magnitude, so the estimated knee point is consistently skewed toward one objective rather than centered, and this skew grows more pronounced as the magnitude gap between the two objectives widens with problem size.
\begin{table}[h]
\centering
\scriptsize
\caption{Knee point preference $\boldsymbol\lambda^\dagger$ for each problem and problem size.}
\label{tab:knee_points}
\begin{tabular}{lcccc}
\toprule
Problem & $n=20$ & $n=50$ & $n=100$ & $n=200$ \\
\midrule
Bi-TSP  & (0.503,0.497) & (0.500,0.500) & (0.501,0.499) & -- \\
Bi-CVRP    & (0.157,0.843) & (0.133,0.867) &(0.107,0.893) & -- \\
Bi-KP      & -- & (0.503,0.497) & (0.492,0.508) & (0.501,0.499) \\
Tri-TSP &(0.335,0.329,0.336) & (0.324,0.338,0.338) &  (0.330,0.338,0.332) & -- \\
\bottomrule
\end{tabular}
\end{table}

\subsection{Anchor placement for three objectives}\label{app:anchor_3obj}
Locating the knee point requires no change when $m=3$. Each extreme preference reduces the problem
to a single-objective TSP on that objective's own coordinates, which is again solved with LKH-3,
and the three resulting extreme points fill the $3\times3$ payoff matrix $\Phi$. 

Placing the remaining anchors, however, has to account for the preference space being a triangle
rather than a segment. Its vertices are the three extreme preferences, and each of its edges is
itself a bi-objective preference space, so it has its own knee: restricting $\mathbf{f}(x^{(i)})$
and $\mathbf{f}(x^{(j)})$ to their $i$-th and $j$-th coordinates gives a $2\times2$ payoff matrix,
to which the same system applies, and we call its solution $\boldsymbol\lambda^\dagger_{ij}$ an
\emph{edge knee}. The extreme points are already available, so the three edge knees cost no
additional solving.

\begin{figure}[H]
    \centering
    \includegraphics[width=0.45\textwidth]{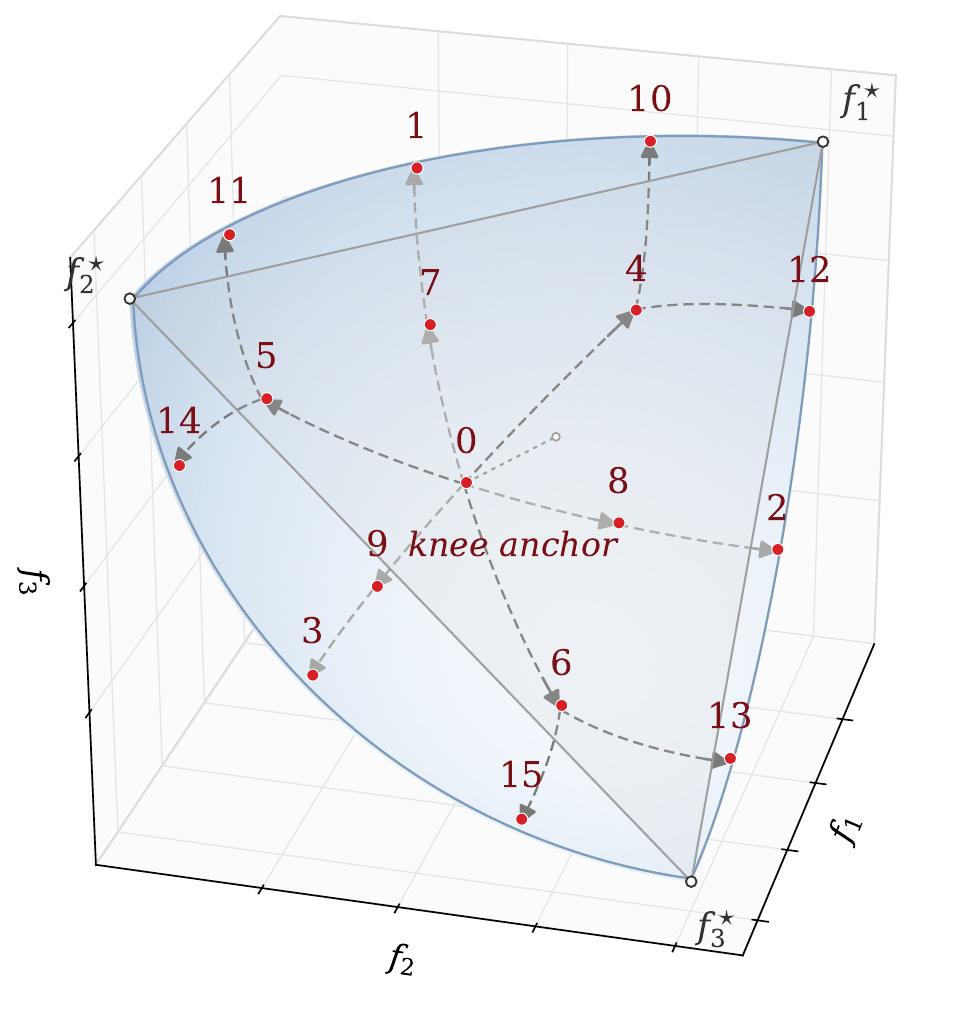}
       \caption{Anchor placement on the three-objective preference simplex with
    $R_1=R_2=R_3=2$, giving $M=16$. The global knee, the three edge knees and the anchors
    placed on the segments they induce are shown together with the transfer chains along
    which the anchors are trained.}
    \label{fig:anchor16}
\end{figure}

These four knees induce three groups of segments, and each segment is divided into equal parts with
one anchor placed at every interior division point, which is the bi-objective rule of
Section~\ref{sec:stage1} applied along one direction at a time. The segments from $\boldsymbol\lambda^\dagger$ to
the three extreme preferences are divided into $R_1$ parts, those from $\boldsymbol\lambda^\dagger$
to the three edge knees into $R_2$ parts, and, within each edge, the two segments that its edge
knee cuts the edge into are divided into $R_3$ parts. Together with the global knee and the three
edge knees themselves, this gives $M = 4 + 3(R_1-1) + 3(R_2-1) + 6(R_3-1)$ anchors, all of them positioned relative to a knee rather than at a fixed geometric location. The
three resolutions are set independently because the triangle is two-dimensional, so a single one
cannot control the spacing towards the extreme preferences, towards the edge knees and along the
edges at once. We set $R_1=R_2=R_3=2$ for Tri-TSP20 and Tri-TSP50, which gives $M=16$, and
$R_1=R_2=R_3=3$ for Tri-TSP100, which gives $M=28$. Figure~\ref{fig:anchor16} shows the resulting
layout for $M=16$.

Transfer training then proceeds outwards from the global knee, the only anchor trained from a
random initialization, along two chains that never meet. On one chain, the anchors on a segment
towards an extreme preference are trained in order of increasing distance from the knee, and the
last of them passes its parameters on to the anchors lying on the two edge segments that end at the
same vertex. On the other, the anchors on a segment towards an edge knee are trained in the same
order, and the last of them passes its parameters on to that edge knee. Every remaining anchor is
therefore adapted for $E_1$ epochs from its nearest already-trained neighbor, as in the
bi-objective case, and anchors at the same distance from the knee are trained in parallel.

\section{Generalization experiments} \label{app:Generalization experiments}
 We further examine how well the solvers of ENS transfer beyond the training setting through two experiments. The first tests generalization to larger sizes: all methods are trained on Bi-TSP100 and evaluated, without any retraining, on Bi-TSP150 and Bi-TSP200. The second replaces the synthetic instances by the real-world KroAB100/150/200 benchmark instances of TSPLIB, whose node coordinates follow a different distribution from the uniformly sampled ones used for training. In both experiments, ENS reuses the anchors and the evolved solvers obtained on Bi-TSP100, so nothing about the model is adapted to the new instances.

Table~\ref{tab:generalization_results} reports the results on the unseen larger sizes. Although its solver set is reused without any adaptation, ENS leads every multi-model baseline at both sizes, with and without augmentation, and stays within $0.07\%$ of the best neural method on Bi-TSP150 and within $0.22\%$ on Bi-TSP200, where the closest competitors are the augmented single-model methods WE-CA and CNH. A set of solvers evolved at $n=100$ therefore remains competitive with the strongest neural methods at sizes it has never seen.

Table~\ref{tab:kro_results} reports the results on the KroAB benchmark, whose node coordinates follow a different distribution from the training instances. On KroAB100, ENS-Aug attains the best hypervolume among all learning-based methods, indicating that ENS is robust to a shift in instance distribution at the size it was evolved for. On KroAB150 and KroAB200, where a size shift is added, ENS-Aug stays within $0.2\%$ of the best neural method and ahead of every multi-model baseline, so the evolved solvers remain competitive even under both kinds of shift at once.

\begin{table}[H]
\centering
\caption{Out-of-distribution generalization results on unseen larger sizes (Bi-TSP150 and Bi-TSP200), with all methods trained on Bi-TSP100.}
\label{tab:generalization_results}
\scriptsize
\setlength{\tabcolsep}{2.8pt}
\renewcommand{\arraystretch}{0.95}
\begin{tabular}{c|lcccccc}
\toprule
& & \multicolumn{3}{c}{Bi-TSP150} & \multicolumn{3}{c}{Bi-TSP200} \\
\cmidrule(lr){3-5} \cmidrule(lr){6-8}
& Method & HV$\uparrow$ & Gap$\downarrow$ & Time$\downarrow$ & HV$\uparrow$ & Gap$\downarrow$ & Time$\downarrow$ \\
\midrule
\multirow{5}{2.3cm}{\centering\scriptsize Non-learnable}
& WS-LKH   & \textbf{0.7148} & \textbf{-1.77\%} & 13h  & \textbf{0.7490} & \textbf{-2.23\%} & 22h  \\
& MOEA/D   & 0.6809 & 3.06\% & 2.4h & 0.7139 & 2.57\% & 2.7h \\
& NSGA-II  & 0.6659 & 5.20\% & 6.8h & 0.7045 & 3.85\% & 6.9h \\
& MOGLS    & 0.6768 & 3.64\% & 22h  & 0.7114 & 2.91\% & 38h  \\
& PPLS/D-C & 0.6784 & 3.42\% & 21h  & 0.7106 & 3.02\% & 32h  \\
\midrule
\multirow{4}{2.3cm}{\centering\scriptsize Single-model}
& PMOCO  & 0.6910 & 1.62\% & 50s  & 0.7231 & 1.31\% & 1.5m \\
& MORAM  & 0.6711 & 4.46\% & 4s   & 0.7024 & 4.14\% & 7s   \\
& CNH    & 0.6985 & 0.56\% & 1.1m & 0.7292 & 0.48\% & 1.9m \\
& WE-CA  & 0.6991 & 0.47\% & 56s  & 0.7264 & 0.86\% & 1.6m \\
\midrule
\multirow{4}{2.3cm}{\centering\scriptsize Multi-model}
& DRL-MOA & 0.6901 & 1.75\% & 45s & 0.7219 & 1.47\% & 1.5m \\
& MDRL    & 0.6922 & 1.45\% & 40s & 0.7251 & 1.04\% & 1.4m \\
& EMNH    & 0.6930 & 1.34\% & 40s & 0.7260 & 0.91\% & 1.4m \\
& ENS     & 0.6968 & 0.80\% & 1.8m & 0.7271 & 0.76\% &  4m   \\
\midrule
\multirow{3}{2.3cm}{\centering\scriptsize Single-model-Aug}
& PMOCO-Aug & 0.6967 & 0.81\% & 47m & 0.7283 & 0.60\% & 1.6h \\
& CNH-Aug   & 0.7025 & -0.01\% & 52m & \underline{0.7343} & \underline{-0.22\%} & 1.7h \\
& WE-CA-Aug & \underline{0.7029} & \underline{-0.07\%} & 51m & 0.7307 & 0.27\% & 1.7h \\
\midrule
\multirow{3}{2.3cm}{\centering\scriptsize Multi-model-Aug}
& MDRL-Aug & 0.6976 & 0.68\% & 47m & 0.7299 & 0.38\% & 1.6h \\
& EMNH-Aug & 0.6983 & 0.58\% & 47m & 0.7307 & 0.27\% & 1.6h \\
& ENS-Aug  & 0.7024   & 0.00\% & 2h   &  0.7327    & 0.00\% & 4.2h  \\
\bottomrule
\end{tabular}
\end{table}

\begin{table*}[h]
\centering
\caption{Results on real-world benchmark instances (KroAB100/150/200).}
\label{tab:kro_results}
\scriptsize
\setlength{\tabcolsep}{2.2pt}
\renewcommand{\arraystretch}{0.85}
\begin{tabular*}{\textwidth}{@{\extracolsep{\fill}}c|lccccccccc@{}}
\toprule
& & \multicolumn{3}{c}{KroAB100} & \multicolumn{3}{c}{KroAB150} & \multicolumn{3}{c}{KroAB200} \\
\cmidrule(lr){3-5} \cmidrule(lr){6-8} \cmidrule(lr){9-11}
& Method & HV$\uparrow$ & Gap$\downarrow$ & Time$\downarrow$ & HV$\uparrow$ & Gap$\downarrow$ & Time$\downarrow$ & HV$\uparrow$ & Gap$\downarrow$ & Time$\downarrow$ \\
\midrule
\multirow{5}{2.1cm}{\centering\tiny Non-learnable}
& WS-LKH   & \textbf{0.7022} & \textbf{-0.21\%} & 2.3m & \textbf{0.7017} & \textbf{-1.20\%} & 4.0m & \textbf{0.7430} & \textbf{-1.91\%} & 5.6m \\
& MOEA/D   & 0.6836 & 2.44\%  & 5.8m & 0.6710 & 3.23\%  & 7.1m & 0.7106 & 2.54\%  & 7.3m \\
& NSGA-II  & 0.6676 & 4.72\%  & 7.0m & 0.6552 & 5.51\%  & 7.9m & 0.7011 & 3.84\%  & 8.4m \\
& MOGLS    & 0.6817 & 2.71\%  & 52m  & 0.6671 & 3.79\%  & 1.3h & 0.7083 & 2.85\%  & 1.6h \\
& PPLS/D-C & 0.6785 & 3.17\%  & 38m  & 0.6659 & 3.97\%  & 1.4h & 0.7100 & 2.62\%  & 3.8h \\
\midrule
\multirow{4}{2.1cm}{\centering\tiny Single-model}
& PMOCO  & 0.6878 & 1.84\% & 9s  & 0.6819 & 1.66\% & 17s & 0.7193 & 1.34\%  & 23s \\
& MORAM  & 0.6723 & 4.05\% & 2s  & 0.6603 & 4.77\% & 2s  & 0.6955 & 4.61\%  & 2s  \\
& CNH    & 0.6947 & 0.86\% & 25s & 0.6892 & 0.61\% & 35s & 0.7250 & 0.56\%  & 41s \\
& WE-CA  & 0.6959 & 0.69\% & 9s  & 0.6898 & 0.52\% & 19s & 0.7232 & 0.81\%  & 25s \\
\midrule
\multirow{4}{2.1cm}{\centering\tiny Multi-model}
& DRL-MOA & 0.6903 & 1.48\% & 10s & 0.6794 & 2.02\% & 18s & 0.7185 & 1.45\% & 23s \\
& MDRL    & 0.6881 & 1.80\% & 10s & 0.6831 & 1.49\% & 17s & 0.7209 & 1.12\% & 23s \\
& EMNH    & 0.6900 & 1.53\% & 9s  & 0.6832 & 1.47\% & 16s & 0.7217 & 1.02\% & 23s \\
& ENS     & 0.6951 & 0.80\% & 1s  & 0.6871 & 0.91\% & 1.2s & 0.7232 & 0.81\% & 2s  \\
\midrule
\multirow{3}{2.1cm}{\centering\tiny Single-model-Aug}
& PMOCO-Aug & 0.6937 & 1.00\% & 12s & 0.6886 & 0.69\%  & 19s & 0.7251 & 0.55\%  & 32s \\
& CNH-Aug   & 0.6980 & 0.39\% & 30s & 0.6938 & -0.06\% & 37s & \underline{0.7303} & \underline{-0.16\%} & 54s \\
& WE-CA-Aug & 0.6997 & 0.14\% & 14s & \underline{0.6944} & \underline{-0.14\%} & 23s & 0.7279 & 0.16\%  & 38s \\
\midrule
\multirow{3}{2.1cm}{\centering\tiny Multi-model-Aug}
& MDRL-Aug & 0.6950 & 0.81\% & 13s & 0.6890 & 0.63\% & 19s & 0.7261 & 0.41\% & 28s \\
& EMNH-Aug & 0.6958 & 0.70\% & 12s & 0.6892 & 0.61\% & 18s & 0.7270 & 0.29\% & 27s \\
& ENS-Aug  & \underline{0.7007} & \underline{0.00\%} & 5s  & 0.6934 & 0.00\% & 35s & 0.7291 & 0.00\% & 76s \\
\bottomrule
\end{tabular*}
\end{table*}

\clearpage
\section{Complete Algorithm}\label{app:algorithm}

Algorithm~\ref{alg:ens} summarizes the complete procedure.

\begin{algorithm}[H]
\caption{Evolving Neural Solver (ENS)}
\label{alg:ens}
\begin{algorithmic}[1]
\Require number of anchors $M$; epochs $E_0,E_1$; target preferences $\{\Lambda_c\}_{c=1}^{M}$; generations $G$; offspring size $K$; reseed ratio $r$; mutation scale $\sigma$; step size $\alpha$; EMA rate $\beta$; evaluation instances $U$
\Ensure anchors $\{\mathbf{w}_c\}_{c=1}^{M}$ and their evolved solver sets $\{S_c\}_{c=1}^{M}$
\Statex \textbf{Stage one: knee point initialization and transfer}
\State estimate the extreme points $\{\mathbf{f}(x^{(i)})\}_{i=1}^{m}$ and obtain $\boldsymbol\lambda^\dagger$ from the payoff matrix $\Phi$
\State place the $M$ anchor preferences $\{\boldsymbol\lambda^{(c)}\}$ relative to $\boldsymbol\lambda^\dagger$
\State train the knee anchor from a random initialization for $E_0$ epochs; \quad $\mathcal{A}\gets\{c^\dagger\}$
\While{$|\mathcal{A}| < M$}
  \State warm-start the untrained anchor $c$ closest to $\mathcal{A}$ from its nearest trained anchor, and train it for $E_1$ epochs under $\boldsymbol\lambda^{(c)}$; \quad $\mathcal{A}\gets\mathcal{A}\cup\{c\}$
\EndWhile
\State $\mathbf{w}_c \gets$ flattened $W_q$ of anchor $c$ for $c=1,\dots,M$, with all remaining weights frozen
\Statex \textbf{Stage two: evolutionary extension}
\For{$c = 1$ \textbf{to} $M$}
  \State $S_c \gets \emptyset$; \quad $\boldsymbol\delta_i^{c} \gets \mathbf{0}$ for $i=1,\dots,m$
\EndFor
\For{$g = 1$ \textbf{to} $G$}
  \Statex \hskip\algorithmicindent \textit{Solution generation}
  \For{$c = 1$ \textbf{to} $M$}
    \State $K_r \gets K$ and $K_e \gets 0$ if $g=1$, otherwise $K_r \gets \mathrm{round}(rK)$ and $K_e \gets K-K_r$
    \State generate $K_r$ reseed offspring $\mathbf{w}' = \mathbf{w}_c + \boldsymbol\epsilon$, \ $\boldsymbol\epsilon\sim\mathcal{N}(\mathbf{0},\sigma^2 I)$
    \State generate $K_e$ elite offspring $\mathbf{w}' = \mathbf{w}_{\text{parent}} + \boldsymbol\epsilon + \alpha\,\boldsymbol\delta_i^{c}$, split evenly among $i=1,\dots,m$, with parents drawn cyclically from $S_c$
    \State $P_c \gets S_c \cup \{\text{offspring}\}$
  \EndFor
  \Statex \hskip\algorithmicindent \textit{Solution evaluation}
  \State sample a batch of $U$ instances
  \For{$c = 1$ \textbf{to} $M$, \textbf{in parallel}}
    \For{\textbf{each} candidate $\mathbf{w}'\in P_c$}
      \State reshape $\mathbf{w}'$ into $W_q$, substitute it into the frozen backbone, and roll out $N$ trajectories on each of the $U$ instances
      \State for each $\boldsymbol\lambda\in\Lambda_c$, apply the internal selection to obtain the representative point $\bar{\mathbf{f}}(\mathbf{w}'\mid\boldsymbol\lambda)$
    \EndFor
  \EndFor
  \State pool the representative points of all candidates across the $M$ anchors and mark those that are non-dominated
  \For{$c = 1$ \textbf{to} $M$}
    \State $S_c \gets \{\mathbf{w}'\in P_c :$ at least one representative point of $\mathbf{w}'$ is non-dominated$\}$
  \EndFor
  \Statex \hskip\algorithmicindent \textit{Searching direction update}
  \For{$c = 1$ \textbf{to} $M$}
    \State assign the survivors of anchor $c$ to the groups $\mathcal{B}_1^{c},\dots,\mathcal{B}_m^{c}$
    \State update $\boldsymbol\delta_i^{c}$ by the EMA of the mean perturbation over $\mathcal{B}_i^{c}$, for $i=1,\dots,m$
  \EndFor
\EndFor
\State \textbf{return} $\{S_c\cup\{\mathbf{w}_c\}\}_{c=1}^{M}$
\end{algorithmic}
\end{algorithm}

\end{document}